\pdfoutput=1

\documentclass[11pt]{article}

\usepackage{naacl2021}

\usepackage{times}
\usepackage{latexsym}
\usepackage{amssymb}
\usepackage{pifont}
\usepackage[T1]{fontenc}

\usepackage[utf8]{inputenc}

\usepackage{microtype}

\usepackage{booktabs}
\usepackage{graphicx}
\usepackage{amsmath}
\usepackage{colortbl}
\usepackage{xspace}
\usepackage{tabularx}
\usepackage{subcaption}
\usepackage{soul}

\newcommand{\hlc}[2][yellow]{{%
    \colorlet{foo}{#1}%
    \sethlcolor{foo}\hl{#2}}%
}

\newcommand{\ourdata}{GovReport\xspace}

\title{Efficient Attentions for Long Document Summarization}

\author{Luyang Huang $^{1}$ \quad Shuyang Cao$^{1}$ \quad Nikolaus Parulian$^{2}$ \quad Heng Ji$^{2}$ \quad Lu Wang$^{1}$ \\
  $^{1}${Computer Science and Engineering, University of Michigan, Ann Arbor, MI} \\
  $^{2}${Department of Computer Science, University of Illinois at Urbana-Champaign, IL} \\
  $^{1}${\tt \{lyhuang, caoshuy, wangluxy\}@umich.edu} \\
  $^{2}${\tt \{nnp2, hengji\}@illinois.edu} \\}

\begin{document}
\maketitle
\begin{abstract}
The quadratic computational and memory complexities of large Transformers have limited their scalability for long document summarization.
In this paper, we propose \textbf{\textsc{Hepos}}, a novel \textit{efficient encoder-decoder attention with head-wise positional strides} to effectively pinpoint salient information from the source. 
We further conduct a systematic study of existing efficient self-attentions. Combined with \textsc{Hepos}, we are able to process ten times more tokens than existing models that use full attentions. 
For evaluation, we present a new dataset, \textbf{\textsc{GovReport}}, with significantly longer documents and summaries. 
Results show that our models produce significantly higher ROUGE scores than competitive comparisons, including new state-of-the-art results on PubMed. 
Human evaluation also shows that our models generate more informative summaries with fewer unfaithful errors. 

\end{abstract}

\section{Introduction}

Long documents, such as scientific papers and government reports, often discuss substantial issues at length, and thus are time-consuming to read, let alone to comprehend. Generating abstractive summaries can help readers quickly grasp the main topics, yet prior work has mostly focused on short texts (containing hundreds of words), e.g., news articles~\cite{gehrmann2018bottom,liu-lapata-2019-text,DBLP:journals/corr/abs-1912-08777}. 

\textit{Model training efficiency} and \textit{summary quality} present a pair of challenges for long document summarization.
State-of-the-art systems~\cite{lewis-etal-2020-bart,DBLP:journals/corr/abs-1912-08777} are built upon Transformer~\cite{NIPS2017_3f5ee243}, which uses attentions to compute pairwise relations between tokens. Such framework has quadratic time and memory complexities, and is too costly for long documents~\footnote{For instance, to fine-tune BART on documents of $10$K tokens with a batch size of $1$, $70$GB of memory is needed for encoder attentions, and $8$GB for encoder-decoder attentions.}. 
Solutions have been proposed to reduce the calculation of \textit{encoder self-attentions}~\cite{wang2020linformer,zaheer2020big} by selectively attending to neighboring tokens~\cite{beltagy2020longformer,child2019generating} or relevant words~\cite{Kitaev2020Reformer:,tay2020sparse}. 
Yet, these methods do not apply to \textit{encoder-decoder attentions} in summarization models since they collaborate and dynamically pinpoint salient content in the source as the summary is decoded. 
Truncation is commonly used to circumvent the issue. However, training on curtailed content further aggravates ``hallucination'' in existing abstractive models~\cite{maynez-etal-2020-faithfulness}. 

We argue that summarizing long documents (e.g., with thousands of words or more) requires efficient handling of both types of attentions. To this end, we propose an efficient encoder-decoder attention with \textbf{head-wise positional strides (\textsc{Hepos})}, where the attention heads follow a strided pattern and have varying starting positions. \textsc{Hepos} reduces computational and memory costs while (1) maintaining the power of emphasizing important tokens, and (2) preserving the global context per head. \textsc{Hepos} successfully doubles the processed input sequence size, when combined with any encoder. To the best of our knowledge, we are the first to study efficient encoder-decoder attentions and provide a systematic comparison of diverse encoder attentions for the task of summarization.\footnote{Our code is released at \url{https://github.com/luyang-huang96/LongDocSum}.} 

For evaluation, we collect \textbf{a new large-scale dataset, \textsc{\ourdata}}, consisting of about $19.5$k U.S. government reports with expert-written abstractive summaries.\footnote{\textsc{\ourdata} can be downloaded from \url{https://gov-report-data.github.io}.} 
\textsc{\ourdata} has two important features: (1) It contains significantly longer documents ($9.4$k words) and summaries ($553$ words) than existing datasets, such as PubMed and arXiv~\cite{cohan-etal-2018-discourse} (see Table~\ref{tab:basic_stat}); (2) Salient content is spread throughout the documents, as opposed to cases where summary-worthy words are more heavily concentrated in specific parts of the document.
These properties make \textsc{\ourdata} an important benchmark for producing long document summaries with multiple paragraphs.

We conduct experiments on \textsc{\ourdata} and scientific papers in PubMed and arXiv. 
First, when summarizing documents of the same length, \textit{ \textsc{Hepos} attention yields significantly better ROUGE scores} than a non-trivial comparison that projects attentions into low-rank space~\cite{wang2020linformer}. 
Second, when trained on the same GPU, \textsc{Hepos} attention, combined with sparse encoder attentions, is able to read more than $10$K words and obtains significantly higher ROUGE scores on \textsc{\ourdata} and new state-of-the-art results on PubMed, compared with full encoder-decoder attention models which can process at most $5$K input words. 
Human judges further rate the summaries generated by our models to be \textit{more informative and faithful}.

We further propose \textbf{a new evaluation metric for faithfulness}, inspired by APES~\cite{eyal-etal-2019-question}, a fill-in-the-blank QA metric for summary evaluation. 
With questions generated from references, our metric, APES$_{src}$, compares QA answers by reading the source and the system summary. It is shown to be better correlated with human judgment than the original metric and an entailment-based scorer~\cite{kryscinski-etal-2020-evaluating}. 

The rest of the paper is organized as follows. We describe efficient encoder attentions in prior work in \S~\ref{sec:encoder}, and formulate our proposed encoder-decoder attention in \S~\ref{sec:decoder}. 
The \textsc{\ourdata} data is presented in \S~\ref{sec:dataset}. We then share details on evaluation metrics (\S~\ref{sec:eval}) and experimental results (\S~\ref{sec:results}). Additional related work is listed in \S~\ref{sec:add_related_work}, with conclusion in \S\ref{sec:conclusion}.

\section{Prior Work on Efficient Encoder Attentions}
\label{sec:encoder}

Transformer models are built upon multi-head attentions in multiple layers. The attention is calculated as
$\textrm{Attention} (\mathbf{Q}, \mathbf{K}, \mathbf{V}) = \textrm{softmax} (\frac{\mathbf{Q} \mathbf{K}^T}{\sqrt{d_k}}) \mathbf{V}$, where $\mathbf{Q}$, $\mathbf{K}$, and $\mathbf{V}$ are query, key, and value matrices, each consisting of $n$ vectors for a document with $n$ tokens, thus the quadratic memory footprint. 

Here, we present an overview of representative methods for efficient encoder self-attentions (henceforth ``encoder attentions'') that can be built upon large pre-trained seq2seq models, e.g., BART~\cite{lewis-etal-2020-bart}. We follow the naming convention of \citet{tay2020efficient}, and summarize their \textit{memory complexities} and numbers of \textit{newly learned parameters} in Table~\ref{tab:model}.

\begin{table}[t]
    \centering
    \fontsize{9}{11}\selectfont
    \setlength{\tabcolsep}{1.5mm}
    \begin{tabular}{@{}lcc@{}}
    \toprule
    \textbf{Model} & \textbf{Complexity} & \textbf{\# New Para.}  \\
    \midrule
    
    \textbf{Full} &$\mathcal{O}(n^2)$&--- \\ \hline
    
    \textbf{Encoder Self-attentions} && \\
    \textit{I. Fixed Patterns} && \\
    {Sliding Window}~\shortcite{beltagy2020longformer} &$\mathcal{O}(n w)$&$0$ \\
    {Adaptive Span}~\shortcite{sukhbaatar-etal-2019-adaptive} & $\mathcal{O}(n \hat{w})$&$\mathcal{O}(1)$\\
    {Global Tokens}~\shortcite{beltagy2020longformer} &$\mathcal{O}( 2ng)$&$0$ \\
    {Stride}~\shortcite{child2019generating} &$\mathcal{O}(n^2/s)$&$0$ \\
    {Random}~\shortcite{zaheer2020big} &$\mathcal{O}(nr)$&$0$ \\
    \hline
    
    \textit{II. Low-rank} && \\
    {Linformer}~\shortcite{wang2020linformer} &$\mathcal{O}(nk)$&$\mathcal{O}(n)$ \\
    \hline
    
    \textit{III. Learnable Patterns} && \\
    {LSH}~\shortcite{Kitaev2020Reformer:} &$\mathcal{O}(l n b_l)$&$0$ \\
    {Sinkhorn}~\shortcite{tay2020sparse} &$\mathcal{O}(2 n b_s)$&$0$ \\
    \hline
    
    \textbf{Encoder-decoder Attentions} \\ 
    {Hepos} (ours) & $\mathcal{O}(m n/s_h)$ & 0 \\ 
    {Linformer} &$\mathcal{O}(mk)$&$\mathcal{O}(n)$ \\ 
    \bottomrule
    \end{tabular}
    \caption{
    Summary of efficient Transformer attentions on \textit{memory complexity} and \textit{newly learned parameters} compared with full attentions at each layer. $m$ and $n$ are lengths of the input and the output. See \S~\ref{sec:encoder} and \S~\ref{sec:decoder} for model-specific hyperparameters. 
    }
    \vspace{-2mm}
    \label{tab:model}
\end{table}

\subsection{Fixed Patterns}
\label{subsec:fixpattern}
Fixed patterns are used to limit the scope of attentions. In our experiments, in addition to window-based attentions, we also combine them with global tokens, stride patterns, or random attentions. 

\smallskip
\noindent \textbf{Sliding window attentions}~\cite{beltagy2020longformer} aim to capture the local context, which is critical for language understanding~\cite{j.2018generating,child2019generating}. Concretely, each query token attends to $w/2$ neighboring tokens on both left and right, yielding a memory complexity of $\mathcal{O}(n w)$.

\smallskip
\noindent \textbf{Adaptive span} is proposed by~\newcite{sukhbaatar-etal-2019-adaptive} to learn attention windows at different layers. This is implemented by learning a masking function for each head independently. In practice, the adaptive span attention has a complexity of $\mathcal{O}(n\hat{w})$, where $\hat{w}$ is the maximum values of predicted spans for all heads. Besides, it introduces $\mathcal{O}(1)$ new parameters for learning spans. 

\smallskip
\noindent \textbf{Global tokens}~\cite{beltagy2020longformer} are often added to sliding windows to let pre-selected tokens attend to the full sequence, to build global representations. Importantly, global attention operations are symmetric, i.e., a global token is also attendable to all tokens in the sequence. We select the first $g$ tokens as global tokens, as leading sentences are often important for summarization. Memory complexity is $\mathcal{O}(2 n g)$ due to the symmetric attentions.

\smallskip
\noindent \textbf{Stride patterns} are proposed by~\newcite{child2019generating} to capture long term interactions, where each query attends to every $s$-th token, with $s$ as the stride size. It thus has a complexity of $\mathcal{O}(n^2/s)$.

\smallskip
\noindent \textbf{Random attention} is motivated by the fact that randomly constructed graphs with $\Tilde{\Theta}(n)$ edges can approximate the complete graphs spectrally~\cite{zaheer2020big}. \newcite{zaheer2020big} propose to allow each query to attend to $r$ random keys, resulting in a complexity of $\mathcal{O}(nr)$. 
For efficient implementations, input tokens are first segmented into blocks. Tokens in the same block attend to tokens in another randomly selected block. 

\subsection{Low-rank Methods}
\label{subsec:linformer}
\newcite{wang2020linformer} show that self-attention matrices are low-rank. They propose \textbf{Linformer} that linearly projects key and value matrices into a low-dimensional space, e.g., from $n$ to $k$, to achieve a $\mathcal{O}(nk)$ complexity. It also introduces $\mathcal{O}(n)$ new parameters for projection matrix learning.

\subsection{Learnable Patterns} 
\label{subsec:learnpattern}

Recently, learnable sparse attentions are proposed to better capture both local and global contexts than attentions based on fixed patterns. 

\smallskip
\noindent \textbf{Locality-sensitive hashing (LSH) attentions} use a random-projection hashing function to hash similar queries and keys into the same buckets in $l$ rounds~\cite{Kitaev2020Reformer:}. Attentions are then computed among tokens within each bucket. For bucket size $b_l$, the complexity of LSH attention is $\mathcal{O}(l n b_l)$.

\smallskip
\noindent \textbf{Sinkhorn attentions} first segment a sequence into blocks, which are then arranged by a learned Sinkhorn sorting network~\cite{tay2020sparse}. Given the new permutation, each query attends to $b_s$ tokens within the same block to maintain the local context and another $b_s$ tokens in a neighboring block to capture global interactions. Its complexity is $\mathcal{O}(2 n b_s)$.

\subsection{Other Attentions} 
We also describe several notable methods that are not suitable for our experiments and excluded from this study: 
\textbf{Recurrence} over input segments are tailored for an autoregressive decoder only~\cite{dai-etal-2019-transformer}; 
\textbf{memory} methods use a separate memory module to attend to full sequences~\cite{pmlr-v97-lee19d}, which share a similar theoretical foundation as global tokens; 
and \textbf{kernel} methods over attentions require training models from scratch~\cite{choromanski2020rethinking, katharopoulos2020transformers}.

\section{Encoder-decoder Attention with Head-wise Positional Strides (Hepos)}
\label{sec:decoder}

\begin{figure}[t]
    \centering
    \includegraphics[width=0.9\columnwidth,trim=0.5cm 1.3cm 0cm 0, clip]{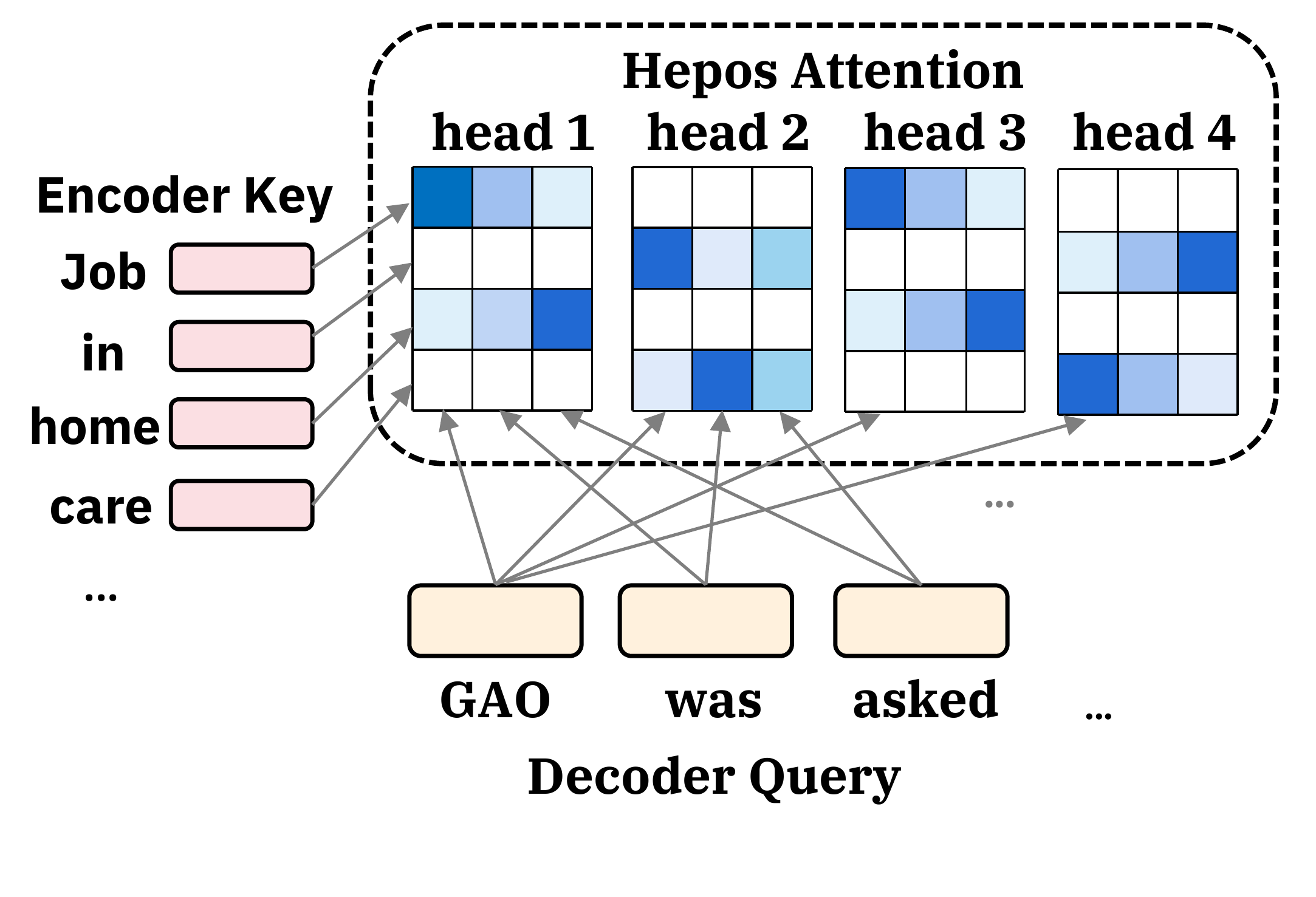}
    \vspace{-2mm}
    \captionof{figure}{
    A toy example of our \textsc{Hepos} attention, with a stride of $2$ and four attention heads. Dark colors indicate that heads 1 and 3 attend to the first and third tokens (``Job" and ``home") in the input, heads 2 and 4 look at the second and fourth words (``in" and ``care"). 
    }
    \label{fig:model}
\end{figure}

The efficient design of encoder-decoder attentions with head-wise positional strides (\textsc{Hepos}) allows models to consume longer sequences. 
Concretely, our design is motivated by two observations: 
(1) Attention heads are redundant~\cite{voita-etal-2019-analyzing}. 
(2) Any individual head rarely attends to several tokens in a row~\cite{clark-etal-2019-bert}. 
Therefore, as illustrated in Fig.~\ref{fig:model}, \textsc{Hepos} uses separate encoder-decoder heads on the same layer to cover different subsets of source tokens at fixed intervals. Each head starts at a different position, and all heads collectively attend to the full sequence. 

Given a stride size of $s_h$, for the $h$-th head, its attention value between decoder query $\mathbf{q}_j$ (at step $j$) and encoder key vector $\mathbf{k}_i$ (for the $i$-th input token) can be formulated as: 

\vspace{-3mm}
{
\fontsize{9}{11}\selectfont
\begin{align}
    a^{h}_{ji} =  \left\{
    \begin{aligned}
    &\mathrm{softmax}(\mathbf{q}_j \mathbf{k}_i),&\ \text{if}\ (i-h)\ \mathrm{mod}\ s_h = 0 \\
    &0 &\text{otherwise}
    \end{aligned}
    \right.
\end{align}
} 

In \textsc{Hepos} attention, each query token attends to $n/s_h$ tokens per head, yielding a memory complexity of $\mathcal{O}(m n/s_h)$, where $m$ is the output length.

For \textbf{comparison}, Linformer (\S~\ref{subsec:linformer}) can be straightforwardly adapted for encoder-decoder attentions by using decoder queries for attention calculation instead. 
We do not adapt pattern-based attentions (\S~\ref{subsec:fixpattern} and \S~\ref{subsec:learnpattern}), since they rely on local token grouping which makes it difficult to pinpoint salient content.

\section{\textsc{GovReport} Dataset}
\label{sec:dataset}

We introduce a new large-scale dataset, \textsc{\ourdata}, containing $19,466$ long reports published by U.S. Government Accountability Office (GAO)\footnote{\url{www.gao.gov}} to fulfill requests by congressional members,
and Congressional Research Service (CRS)\footnote{\url{crsreports.congress.gov}}, covering researches on a broad range of national policy issues. A human-written summary is provided along with each report. During data collection, we remove boilerplates from crawled files, and keep the section and paragraph structure of the documents and summaries. Additional data cleaning and processing details are in Appendix~\ref{sec:append_gov_dataset}. 

We obtain $12,228$ GAO reports and $7,238$ CRS reports of high quality evidenced by human inspection of $200$ parsed reports. Collected GAO reports and CRS reports have on average $6.9$ and $4.6$ sections, respectively. We split train, validation and test set by publication date on each dataset, and end up with $17519$ training samples, $974$ validation documents, and $973$ test samples.

Notably, summaries of GAO reports are written by experts, and are often structured into three aspects in order: ``\texttt{Why GAO did this study}''---motivation and problem(s) under discussion, ``\texttt{What GAO found}''---findings of the report, and ``\texttt{What GAO recommends}''---suggestions and solutions to the problem(s). All but three GAO summaries include ``What GAO Found''. The percentages of GAO summaries that contain ``Why GAO did this study'' and ``What GAO recommends'' are $94.8\%$ and $29.0\%$. 
For comparison, structured summaries are also observed on \textsc{PubMed}~\cite{cohan-etal-2018-discourse} samples. Though they do not contain explicit aspect labels, the summaries can often be broken down into ``Introduction'', ``Methods'', ``Results'', and ``Conclusion'' via keyword matching. Details about keyword choices for each aspect are provided in Table~\ref{table:aspect_pubmed} in Appendix~\ref{sec:append_human_eval}.

\begin{table}[t]
    
    \fontsize{9}{11}\selectfont
    \setlength{\tabcolsep}{1.3pt}
    \hspace{-1mm}
    \begin{tabular}{lcccccc}
    \toprule
        \textbf{Dataset} & \textbf{\# Doc} & \multicolumn{2}{c}{\textbf{Summary}} & \textbf{Doc} & \textbf{Comp.} & \textbf{Den.} \\
        & & \# word & \# sent & \# word \\
        \midrule
        \textsc{PubMed} & 133,215 & 202.4 & 6.8 & 3049.0 & 16.2 & 5.8 \\
        \textsc{arXiv} & 215,913 & 272.7 & 9.6 & 6029.9 & 39.8 & 3.8 \\
        \textsc{BillSum} & 23,455 & 207.7 & 7.2 & 1813.0 & 13.6 & 4.1 \\
        \textsc{BigPatent} & 1,341,362 & 116.5 & 3.7 & 3573.2 & 36.3 & 2.4 \\
        \rowcolor{red!15} \textsc{\ourdata} & 19,466 & \textbf{553.4} & \textbf{17.8} & \textbf{9409.4} & 19.0 & 7.3 \\
        \bottomrule
    \end{tabular}
    \caption{
    Statistics of \textsc{\ourdata} and existing long document summarization datasets. 
    \textbf{Comp.}: compression ratio, \textbf{Den.}: extractive fragment density~\cite{grusky-etal-2018-newsroom}. 
    All values are mean over the whole dataset except for the ``\# Doc'' column. 
    Documents and summaries in \textsc{\ourdata} are significantly longer.}
    \label{tab:basic_stat}
\end{table}

\begin{figure}[t]
    \centering
    \vspace{-2mm}
    \includegraphics[width=0.45\textwidth]{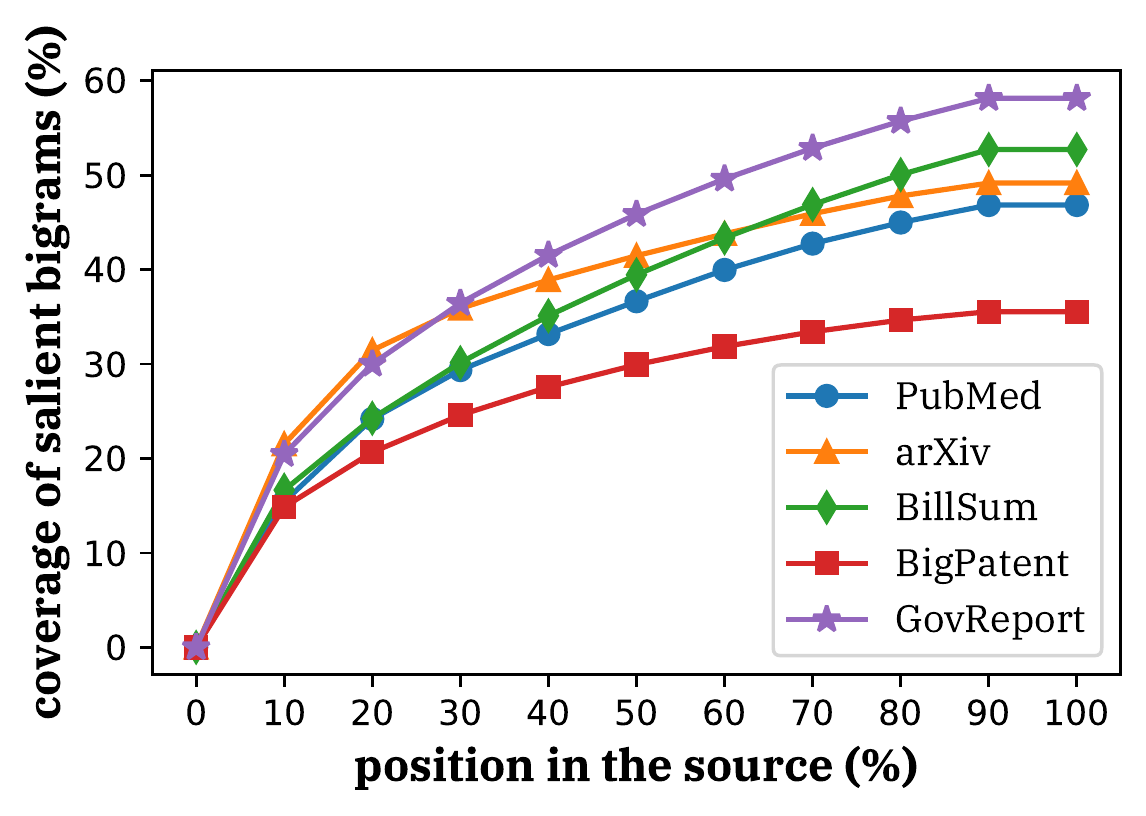}
    \vspace{-3mm}
    \caption{
    Percentage of unique salient bigrams accumulated from the start to X\% of the source. Key information is spread over the documents in \textsc{\ourdata}, highlighting the importance of understanding longer text.}
    
    \label{fig:accumulated_occurrence}
\end{figure}

\paragraph{Comparison with Existing Long Document Summarization Datasets.} 
In Table~\ref{tab:basic_stat}, we compare \textsc{\ourdata} with several existing long document summarization datasets, including \textsc{PubMed} and \textsc{arXiv}~\cite{cohan-etal-2018-discourse} that consist of scientific publications; \textsc{BillSum}~\cite{kornilova-eidelman-2019-billsum}, a collection of congressional bills; and \textsc{BigPatent}~\cite{sharma-etal-2019-bigpatent}, a corpus of U.S. patent documents. 

First, \textit{documents and summaries in \ourdata are significantly longer than prior datasets}. 
Next, we inspect the distribution of summary-worthy bigrams in the source by dividing each document into ten equisized partitions. 
For each partition, we count the occurrence of unique bigrams that also appear in the reference, accumulated from the start of the document to the end of the partition. 
Fig.~\ref{fig:accumulated_occurrence} shows that \textit{key information is spread throughout documents in \textsc{\ourdata}},
with new salient bigrams being steadily added as more content is consumed.
For \textsc{arXiv} and \textsc{BigPatent}, only about $10\%$ of new salient bigrams are accumulated in the second half of the documents, reflecting the heavy positional bias in these two datasets. 
In contrast, in \ourdata and \textsc{BillSum}, more than $18\%$ of new summary-worthy bigrams appear in the later half of the articles, showing a more even distribution. A similar trend is observed on unigrams. 
However, \textsc{BillSum} has the shortest documents among the five datasets. 

\section{Summary Evaluation with Cloze QA}
\label{sec:eval}

This work aims to evaluate whether processing more text improves both informativeness and faithfulness of abstractive summaries. In addition to ROUGE~\cite{lin-2004-rouge} and human evaluation, we extend existing QA-based metric~\cite{eyal-etal-2019-question} and consider an entailment-based scorer.

\smallskip
\noindent \textbf{QA-based Evaluation.} 
We present a new faithfulness evaluation metric by extending the \textbf{APES} score~\cite{eyal-etal-2019-question}. 
We follow APES to construct a set of \textbf{cloze questions}, $\{q\}$, from each reference summary by masking entities.
Events, dates, and numbers are also masked, as they are prevalent in our data. Each masked phrase becomes the gold-standard answer $a_{ref}$ for a question $q$. 
We do not generate natural language questions~\cite{durmus-etal-2020-feqa, wang-etal-2020-asking}, due to the lack of accurate question generation models for the domains of government reports and scientific papers.

QA models are trained by reading a question and a \textbf{context} to label the answer span in the context. We construct context by greedily selecting sentences that maximize the improvement of ROUGE-2 recall when compared with the reference summary. If the answer $a_{ref}$ cannot be found in the context, the sample is excluded from training. We train all QA models by fine-tuning BERT~\cite{devlin-etal-2019-bert} to predict the answer span.

To evaluate the faithfulness of a system summary, \textbf{\textsc{APES}} uses the QA model to read the summary and a question $q$ to label an answer $a_{sys}$. It calculates a unigram F1 score by \textit{comparing $a_{sys}$ and $a_{ref}$}. 
Different from \textsc{APES}, we further use the QA model to read the context (sentences selected from the source) and give an answer $a_{cxt}$ to the question $q$. We compute a unigram F1 by \textit{comparing $a_{sys}$ and $a_{cxt}$}, denoted as \textbf{\textsc{APES}$_{src}$}. Given that existing summarization models rarely rewrite names or numbers correctly, our metric can better capture faithfulness by using a gold-standard answer constructed from the source article than from the human-written abstract. 

To \textbf{extract entities and events}, we deploy a state-of-the-art IE framework, OneIE~\cite{lin2020oneie} on \textsc{\ourdata}.
On PubMed, we re-train OneIE on Genia 2011~\cite{genia2011} and 2013~\cite{genia2013}, and PubMed~\cite{wei2019pubtator} datasets to extract domain-specific entities and events, such as entities of \textit{Gene} and \textit{Disease}. 
We additionally include numbers and dates extracted by spaCy~\cite{spacy2}.

\smallskip
\noindent \textbf{Entailment-based Evaluation.} We further consider \textbf{FactCC}~\cite{kryscinski-etal-2020-evaluating}, which evaluates factual consistency of a system summary by predicting an entailment score between the source and the summary. We reproduce their method on our datasets. 

Additional details for implementing the evaluation models and the entity extraction models are given in Appendix~\ref{sec:append_eval_detail}.

\section{Experimental Results}
\label{sec:results}

In this section, we start with describing training details in \S~\ref{subsec:training_details}. 
We then compare attention variants on documents of the same length (\S~\ref{subsec:length_control}) and study whether reading more text can generate more informative summaries (\S~\ref{subsec:memory_control}). 
We further report human evaluation on summary informativeness and faithfulness as well as automatic faithfulness scores (\S~\ref{subsec:humaneval}). Finally, we investigate whether automatic metrics correlate with human judgment (\S~\ref{subsec:autoeval}).

\subsection{Training Details} 
\label{subsec:training_details}

We fine-tune BART~\cite{lewis-etal-2020-bart} for all experiments. We implement our models with PyTorch~\cite{NEURIPS2019_9015} and Fairseq~\cite{ott2019fairseq}. Additional position embeddings are initialized randomly for models that handle longer inputs. 
The learning rate is set to $1 \times 10^{-4}$ and learning rate warm-up is applied for the first 10,000 steps. Adafactor~\cite{pmlr-v80-shazeer18a} optimizer with a gradient clipping of 0.1 is used. 
All models are trained on two Quadro RTX 6000 GPUs with 24GB memory or one Quadro RTX 8000 with 48GB memory. We set a batch size of 2 per step and accumulate gradient every 32 steps. 
During test, we adopt a beam size of 4 and a length penalty of 2~\cite{wu2016google} on all datasets.

\subsection{Comparing Attention Variants}
\label{subsec:length_control}

\noindent \textbf{Comparisons.} 
We first experiment with articles that are all truncated at $1024$ tokens. 
For encoder attentions, we consider the following variants: 
(1) sliding \textsc{Window}; 
(2) adaptive span (\textsc{AdaSpan}); 
(3) \textsc{Global} tokens; 
(4) \textsc{Stride}; 
(5) \textsc{Random} tokens; 
(6) Linformer (\textsc{Lin.}); 
(7) locality sensitive hashing (\textsc{LSH}); 
and (8) \textsc{Sinkhorn}. 
We ensure models are comparable by setting hyperparameters to satisfy $w=\hat{w}=k=l b_{l}=2b_{s}=256$, so that models have similar memory complexity. 
For LSH attentions, we select $l=4$ rounds of hashing. 
Following prior work~\cite{zaheer2020big}, we combine \textsc{Global}, \textsc{Stride}, and \textsc{Random} with \textsc{Window} and \textsc{AdaSpan}, where we set $g=n^{2}/s=r=128$ for a fair comparison. 
We adapt Linformer to encoder-decoder attentions to compare with \textsc{Hepos}, where we use $s_h=n/k=4$ for all experiments. 
Finally, we report results using \textsc{FULL}, i.e., the original, encoder and encoder-decoder attentions.

\begin{table}[t]
    \centering
    \fontsize{9}{11}\selectfont
    \setlength{\tabcolsep}{0.1mm}
    \begin{tabular}{lcccccc}
    \toprule
    {} & \multicolumn{3}{c}{\textbf{GovReport (new)}} & \multicolumn{3}{c}{\textbf{PubMed}} \\
    System & R-1 & R-2 & R-L & R-1 & R-2 & R-L \\
    \midrule
       \textsc{FULL} & 52.83 & 20.50 & 50.14  & 45.36 & 18.74 & 40.26  \\
    \hline
    \multicolumn{7}{l}{{\bf Encoder variants w/ full enc-dec attn.}}\\
    \multicolumn{2}{l}{\textit{I. Fixed Patterns}} && &&&\\
      \textsc{Window} & 50.78 & 18.59 & 48.10 & 42.74 & 16.83 & 37.96  \\
      $+$ \textsc{Global} & 51.24 & 19.01 & 48.58 & 43.44 & 17.07 & 38.55  \\
      $+$ \textsc{Stride} & 51.53 & 19.14 & 48.68 & \colorbox{red!35}{{43.73}} & \colorbox{red!35}{\makebox[2em]{{17.25}}} & \colorbox{red!35}{\makebox[2em]{{38.82}}}  \\
      $+$ \textsc{Random} & 51.49 & 18.90 & \colorbox{red!35}{\makebox[2em]{{48.75}}} & 43.38 & 16.87 & 38.45   \\
      \textsc{AdaSpan} & 50.76 & 18.69 & 48.13 & 43.42 & 17.16 & 38.60  \\
      $+$ \textsc{Global} & 50.33 & 18.56 & 47.80 & 43.24 & 17.01 & 38.42  \\
      $+$ \textsc{Stride} & \colorbox{red!35}{\makebox[2em]{{51.56}}} & \colorbox{red!35}{\makebox[2em]{{19.19}}} & 48.57 & 43.71 & \colorbox{red!35}{\makebox[2em]{{17.25}}} & 38.76 \\
      $+$ \textsc{Random} & 51.39 & 18.89 & 48.74 & 43.28 & 16.87 & 38.45  \\
      \multicolumn{2}{l}{\textit{II. Low-Rank Methods}} && \\
      \textsc{Lin.} & 50.70 & 18.48 & 47.85  & 43.65 & 17.12 & 38.71 \\
      \multicolumn{2}{l}{\textit{III. Learnable Patterns}} &&& \\
      \textsc{LSH} & 51.95 & 19.36 & 48.85  & 44.74 & 18.07 & 39.76  \\
      \textsc{Sinkhorn} & \colorbox{orange!45}{\makebox[2em]{53.00$^\ast$}} & \colorbox{orange!45}{\makebox[2em]{20.05$^\ast$}} & \colorbox{orange!45}{\makebox[2em]{50.25$^\ast$}} & \colorbox{orange!45}{\makebox[2em]{45.10}} & \colorbox{orange!45}{\makebox[2em]{18.40$^\ast$}} & \colorbox{orange!45}{\makebox[2em]{40.11$^\ast$}}   \\
    \hline
    \multicolumn{7}{l}{{\bf Enc-dec variants w/ {full} encoder attn. }} \\
    \textsc{Lin.} & 47.79 & 14.93 & 45.15 &  45.16 & 17.66 & 40.25    \\
    \textsc{Hepos} (ours) & \colorbox{blue!25}{\makebox[2em]{{51.05}$^\ast$}} & \colorbox{blue!25}{\makebox[2em]{19.44$^\ast$}} & \colorbox{blue!25}{\makebox[2em]{{48.51}$^\ast$}} & \colorbox{blue!25}{\makebox[2em]{45.80}$^\ast$} & \colorbox{blue!25}{\makebox[2em]{18.61}$^\ast$} & \colorbox{blue!25}{\makebox[2em]{40.69}$^\ast$}   \\
    \multicolumn{7}{l}{{\bf Enc-dec variants w/ {Sinkhorn} encoder attn.}} \\
    \textsc{Lin.} & 42.90 & 12.86 & 40.32 & 44.84 & 17.65 & \colorbox{blue!25}{\makebox[2em]{{39.98}}}    \\
    \textsc{Hepos} (ours) & \colorbox{blue!25}{\makebox[2em]{51.34$^\ast$}} & \colorbox{blue!25}{\makebox[2em]{{19.09}$^\ast$}} & \colorbox{blue!25}{\makebox[2em]{48.73$^\ast$}}  & \colorbox{blue!25}{\makebox[2em]{{44.85}}} & \colorbox{blue!25}{\makebox[2em]{{18.19}$^\ast$}} & 39.91    \\
    \bottomrule
    \end{tabular}
    \caption{
    Results on evaluating encoder and encoder-decoder attentions on input of the same length. Best ROUGE scores of \textit{fixed patterns}, \textit{learnable patterns}, and \textit{enc-dec attentions} are in \hlc[red!35]{red}, \hlc[orange!45]{orange}, and \hlc[blue!25]{purple}, respectively. $\ast$: significantly better than comparison(s) using the same encoder or enc-dec attention (approximation randomization test, $p<0.0005$).}
    \label{tab:length_control}
\end{table}

\smallskip
\noindent \textbf{Results.} 
\textit{Among all \textbf{encoder} variants, learnable patterns perform the best, approaching the performance of full attentions} on both \ourdata and PubMed, as shown in Table~\ref{tab:length_control}. Within learnable patterns, Sinkhorn attention consistently obtains better ROUGE scores. 
Moreover, combining techniques in fixed patterns is more effective than simply using window-based sparse attentions, though with an increased memory cost. 

For \textit{\textbf{encoder-decoder} attentions, \textsc{Hepos} consistently yields higher ROUGE scores than Linformer on both datasets}, using either full or Sinkhorn encoder. 
Notably, coupled with a Sinkhorn attention, our model's performance matches the variant using full encoder attention, implying the effectiveness of \textsc{Hepos} on both identifying the salient content and capturing the global context. 

\subsection{Reading More Input Boosts Informativeness}
\label{subsec:memory_control}

\begin{table}[t]
    
    \fontsize{9}{11}\selectfont
    \setlength{\tabcolsep}{0.7mm}
    \hspace{-2mm}
    \begin{tabular}{lcccccc}
    \toprule
    {} & \multicolumn{3}{c}{\bf GovReport} & \multicolumn{3}{c}{\bf PubMed}   \\
    System (\textsc{MaxLen}) & R-1 & R-2 & R-L & R-1 & R-2 & R-L \\
    \midrule
    \multicolumn{7}{l}{\textbf{Baselines}} \\
       \textsc{PEGASUS} (1024)  & -- & -- & --  & 45.97 & 20.15 & 41.34  \\
       \textsc{TLM} (full) & -- & -- & -- & 42.13 & 16.27 & 39.21   \\
       \textsc{SEAL} (full) & -- & -- & --  & 46.50 & 20.10 & 42.20  \\
       \textsc{DANCER} (full)  & -- & -- & --  & 46.34 & 19.97 & 42.42 \\
       \textsc{BigBird} (3072) & -- & -- & -- & 46.32 & 20.65 & 42.33  \\

    \hline
    \multicolumn{7}{l}{\textbf{Encoder variants w/ full enc-dec attn.}} \\
    \textsc{Full} (1024) & 52.83 & 20.50 & 50.14 & 45.36 & 18.74 & 40.26  \\
    \textsc{Stride} (4096) & 54.29 & 20.80 & 51.35 & 46.95 & 19.98 & 41.67  \\
    \textsc{Lin.} (3072) & 44.84 & 13.87 & 41.94 & 43.69 & 16.35 & 38.66 \\
    \textsc{LSH} (4096) & 54.75 & 21.36 & 51.27 & 47.54 & 20.79 & 42.22  \\
    \textsc{Sinkhorn} (5120) & 55.45 & 21.45 & 52.48 & 47.96 & 20.78 & 42.53  \\
    \midrule
    \multicolumn{7}{l}{\textbf{Encoder variants w/ \textsc{Hepos} enc-dec attn.} (ours)} \\
    \textsc{LSH} (7168) & 55.00 & 21.13 & 51.67 & \textbf{48.12} & \textbf{21.06} & \textbf{42.72}  \\
    \textsc{Sinkhorn} (10240) & \textbf{56.86} & \textbf{22.62} & \textbf{53.82} & 47.93 & 20.74 & 42.58  \\

    \bottomrule
    \end{tabular}
    \caption{
    ROUGE scores for models trained on the same GPU. 
    \textsc{Sinkhorn} with \textsc{Hepos} enc-dec attention and \textsc{LSH} with \textsc{Hepos} both read more text and obtain significantly better scores than other models on GovReport and PubMed ($p<0.0005$). 
    }
    \label{tab:rouge_memory_control}
\end{table}

\begin{table}[t]
    \centering
    \fontsize{9}{11}\selectfont
    \setlength{\tabcolsep}{3mm}
    \begin{tabular}{lccc}
    \toprule
    System (\textsc{MaxLen}) & R-1 & R-2 & R-L  \\
    \midrule
    \multicolumn{4}{l}{\textbf{Baselines}} \\
    \textsc{PEGASUS} (1024)  & 44.21 & 16.95 & 38.83  \\
     \textsc{TLM} (full) & 41.62 & 14.69 & 38.03   \\
     \textsc{SEAL} (full) & 44.3 & 18.0 & 39.3    \\
     \textsc{DANCER} (full)  & 45.01 & 17.60 & 40.56   \\
     \textsc{BigBird} (3072) & 46.63 & 19.02 & 41.77 \\
    \hline
    \multicolumn{4}{l}{\textbf{Encoder variants w/ \textsc{Hepos} enc-dec attn.} (ours)} \\
    \textsc{LSH} (7168) & \textbf{48.24} & \textbf{20.26} & \textbf{41.78} \\
    \textsc{Sinkhorn} (10240) & 47.87 & 20.00 & 41.50  \\
    \bottomrule
    \end{tabular}
    \caption{
    Automatic evaluation on arXiv. Our best model yields better ROUGE scores than previous state-of-the-art models.}
    \label{tab:arxiv}
\end{table}

We investigate whether processing more words generates more informative summaries. 

\smallskip
\noindent \textbf{Comparisons} include recent top-performing \textit{abstractive} models:  \textsc{PEGASUS}~\cite{DBLP:journals/corr/abs-1912-08777}, a large pre-trained summarization model with truncated inputs;  \textsc{TLM}~\cite{pilault-etal-2020-extractive},  \textsc{DANCER}~\cite{Gidiotis2020ADA}, and \textsc{SEAL}~\cite{zhao2020seal}, all of them using hybrid extract-then-abstract methods; and  \textsc{BigBird}~\cite{zaheer2020big}, which combines sliding window, global and random token attentions in the encoder. 

For encoder variants, we pick the best performing model from fixed patterns to be combined with full encoder-decoder attention, i.e., sliding window with stride (\textsc{Stride}), low-rank method (\textsc{Lin.}), and learnable patterns (\textsc{LSH} and \textsc{Sinkhorm}). 
We then combine learnable patterns with \textsc{Hepos} to support processing more text. 
All models consume as long an input as the memory allows.

\smallskip
\noindent \textbf{Results.} 
Overall, \textit{models that read more text obtain higher ROUGE scores}, according to results on \ourdata and PubMed in Table~\ref{tab:rouge_memory_control}. 
First, different encoder variants with full encoder-decoder attentions attain better results than the full attentions baseline except Linformer. 
Second, \textit{adding \textsc{Hepos} encoder-decoder attention almost doubles the words that can be processed and further improves the performance.} 
This highlights the importance of handling both encoder attentions and encoder-decoder attentions efficiently. 
Notably, \textsc{Hepos} with an LSH encoder \textit{achieves new state-of-the-art results on PubMed}, outperforming BigBird which only uses sparse attentions on the encoder. 
We also report performances of our two best models with \textsc{Hepos} on arXiv in Table~\ref{tab:arxiv}, and they outperform all competitive abstractive models.

\begin{figure}[t]
    \centering
    \fontsize{9}{10}\selectfont
    \setlength{\tabcolsep}{1.5mm}{
    \begin{tabular}{p{74mm}}
    \toprule
    \textbf{Human-written Summary:} \\
    In fiscal year 2018, Medicaid covered approximately 75 million individuals at an estimated cost of \$629 billion, \$393 billion of which were federal funds.
    (...) \\
    While CMS is generally required to disallow, or recoup, federal funds from states for eligibility-related improper payments if the state's eligibility error rate exceeds 3 percent, it has not done so for decades, because \textcolor{blue!75}{\textbf{the method it used for calculating eligibility error rates was found to be insufficient for that purpose.}} To address this, in July 2017, CMS \textcolor{blue!75}{\textbf{issued revised procedures through which it can recoup funds for eligibility errors}}, beginning in fiscal year 2022.  (...) \\
    \hline
    \textbf{Model w/ full attn.:} \\
    Medicaid is a federal-state program that provides health care coverage to low-income individuals and families.
    (...) \\
    CMS officials stated that \textcolor{red!55}{\textbf{they have provided states with guidance}} on how to use data from SSA's automated system for eligibility determinations, (...) \\
    CMS officials said that \textcolor{red!55}{\textbf{they did not have guidance}} on when states should use SSA data to evaluate eligibility based on nonfinancial or financial criteria. (...) \\
    \hline
    \textbf{Model w/ \textsc{Hepos} enc-dec attn.} (ours): \\
    The Patient Protection and Affordable Care Act (PPACA) expanded Medicaid coverage to millions of low-income adults and children with disabilities and their eligible dependents.
    (...) \\
    The selected states also reported that \textcolor{blue!75}{\textbf{they did not have adequate processes to address these issues.}} CMS has taken steps to \textcolor{blue!75}{\textbf{improve its oversight of the Medicaid program, including issuing guidance to states}} on the use of MAGI-exempt bases for determining eligibility, but these efforts have not been fully implemented.
    (...) \\
    \bottomrule
    \end{tabular}
    }
    \caption{Sample summaries for a government report. The model with truncated input generates \textcolor{red!55}{\textbf{unfaithful content}}.
    \textsc{Hepos} attention with a Sinkhorn encoder covers \textcolor{blue!75}{\textbf{more salient information}}.}
    \label{fig:sample_outputs}
\end{figure}
\begin{figure}[!h]
    \centering
    \includegraphics[width=0.8\columnwidth]{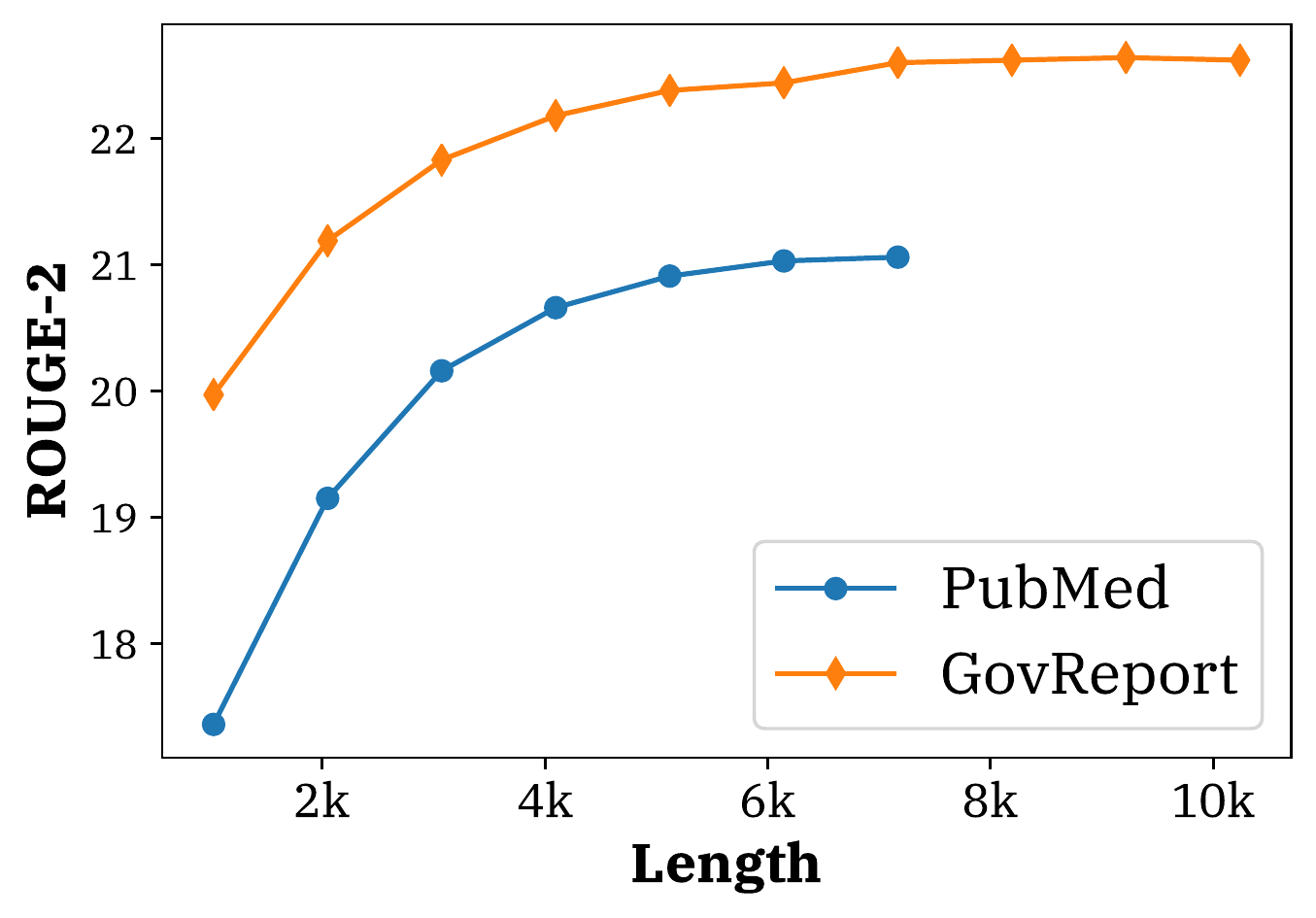}
    \captionof{figure}{
    Summarizing articles truncated at different lengths by the best models: \textsc{LSH} (7168)$+$\textsc{Hepos} on PubMed and \textsc{Sinkhorn} (10240)$+$\textsc{Hepos} on GovReport. Reading more consistently improves ROUGE-2.
    }
    \label{fig:length_effect}
\end{figure}

As can be seen from the sample summaries in Fig.~\ref{fig:sample_outputs}, our model that reads in $10$k tokens generates more informative summary than the full attention model that only processes $1$k tokens.
Fig.~\ref{fig:length_effect} further shows that ROUGE-2 scores can be consistently lifted when reading more input, with similar trends observed on ROUGE-1 and ROUGE-L. 
More sample outputs are presented in Appendix~\ref{sec:append_sample_output}.

\subsection{Reading More Input Improves Faithfulness}
\label{subsec:humaneval}
Here we first show \textbf{human evaluation} results on informativeness and unfaithful errors in the generated summaries. 
We sample 100 documents from \ourdata and PubMed (50 each) with structured references that are labeled with aspects as described in \S~\ref{sec:dataset} and Appendix~\ref{sec:append_human_eval}. 
Each sample is evaluated by two fluent English speakers, who have cumulatively annotated tens of thousands of sentences for the same tasks before this work. 
Annotators are asked to label each summary sentence with an aspect and then decide whether it contains any type of error. 
Three types of unfaithful errors are considered: (i) \textbf{hallucination}---fabricating content not present in the input, (ii) \textbf{deletion}---incorrectly deleting crucial entities, events, or clauses, and (iii) \textbf{false concatenation}---inappropriately concatenating components from different sentences. $1$ is given if any judge determines that a certain type of error exists in the sentence, $0$ otherwise. 

After reading the full summaries, each judge also scores aspect-level \textbf{informativeness}---whether the summary covers important information of an aspect when compared with the reference. All system summaries and references are presented in a random order. 
Human evaluation guidelines and sample summaries for different aspects are included in Appendix~\ref{sec:append_human_eval}.

\begin{table}[t]
    
    \fontsize{9}{11}\selectfont
    \setlength{\tabcolsep}{1.5mm}
    \begin{tabular}{lccccc}
    \toprule
    System (MaxLen) & \textbf{Inf.}$\uparrow$ &  \textbf{Hal.}$\downarrow$  &  \textbf{Del.}$\downarrow$ & \textbf{Concat.}$\downarrow$ \\
    \midrule
    \multicolumn{2}{l}{\textit{GovReport}} &&&\\
    \multicolumn{5}{l}{\textbf{Encoder variants w/ full enc-dec attn.}} \\
    \textsc{Full} (1024) & 3.29 & 15.2\% & 3.5\% & 9.5\% \\
     \textsc{Sinkhorn} (5120) & 3.32 & \textbf{11.0\%} & \textbf{2.3\%} & 9.4\%\\
     \multicolumn{5}{l}{\textbf{Encoder variant w/ \textsc{Hepos} enc-dec attn.} (ours)} \\
     \textsc{Sinkhorn} (10240) & \textbf{3.53} & 11.5\% & 3.4\% & \textbf{8.8\%} \\
     \midrule
     \multicolumn{2}{l}{\textit{PubMed}} &&&\\
     \multicolumn{5}{l}{\textbf{Encoder variants w/ full enc-dec attn.}} \\
     \textsc{Full} (1024) & 3.27 & 20.1\% & 2.8\% & 14.3\% \\
     \textsc{Sinkhorn} (5120)  & 3.94 & 4.8\% & \textbf{1.6\%} & 9.6\%\\
     \multicolumn{5}{l}{\textbf{Encoder variant w/ \textsc{Hepos} enc-dec attn.} (ours)} \\
     \textsc{Sinkhorn} (10240) & \textbf{4.18} & \textbf{3.5\%} & 2.2\% & \textbf{9.1\%} \\
    \bottomrule
    \end{tabular}
    \caption{Human evaluation on informativeness (Inf.) (1-to-5), and percentages of unfaithful errors due to hallucination (Hal.), deletion (Del.), and false concatenation (Concat.). Inter-rater agreement with Krippendorf's $\alpha$ for all columns: 0.59, 0.59, 0.53 and 0.60. 
    }
    \label{tab:human_eval}
\end{table}

\smallskip
\noindent \textbf{Results.} 
Overall, \textit{reading more text significantly improves informativeness as well as reduces fabricated content.} From Table~\ref{tab:human_eval}, we observe that \textsc{Hepos} attention, combined with a \textsc{Sinkhorn} encoder, obtains better informativeness scores than comparisons that read in less text on both datasets. This echos results from automatic evaluation in the previous section. 
Moreover, both models that use efficient attentions reduce unfaithfulness, especially hallucination errors, when compared with the full attention model, which only reads 1024 tokens. 
As the models read more content, they learn to surface more factual and richer content in the summaries, as seen in Fig.~\ref{fig:sample_outputs}.

\begin{figure}[t]
    \centering
    \includegraphics[width=0.99\columnwidth]{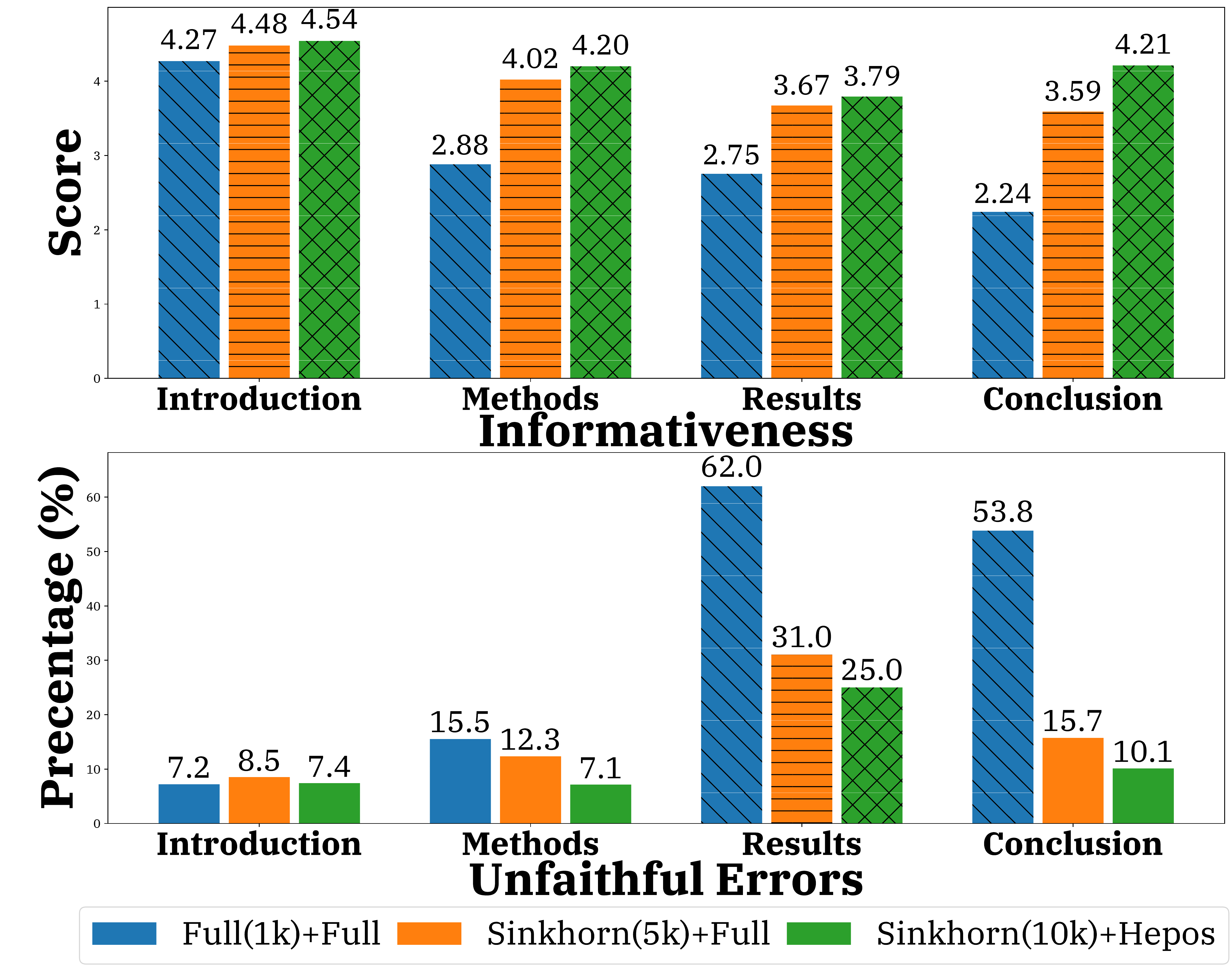}
    \vspace{-5mm}
    \captionof{figure}{Aspect-level informativeness and percentages of sentences containing unfaithful errors as labeled by both human judges on PubMed. Models with efficient attentions reduce errors for later sections in the sources, e.g., ``Results" and ``Conclusion". 
    }
    \label{fig:human_eval_aspect_level_pubmed}
\end{figure}

\begin{figure}[t]
    \centering
    \includegraphics[width=0.99\columnwidth,trim=0cm 0 0.1cm 0, clip]{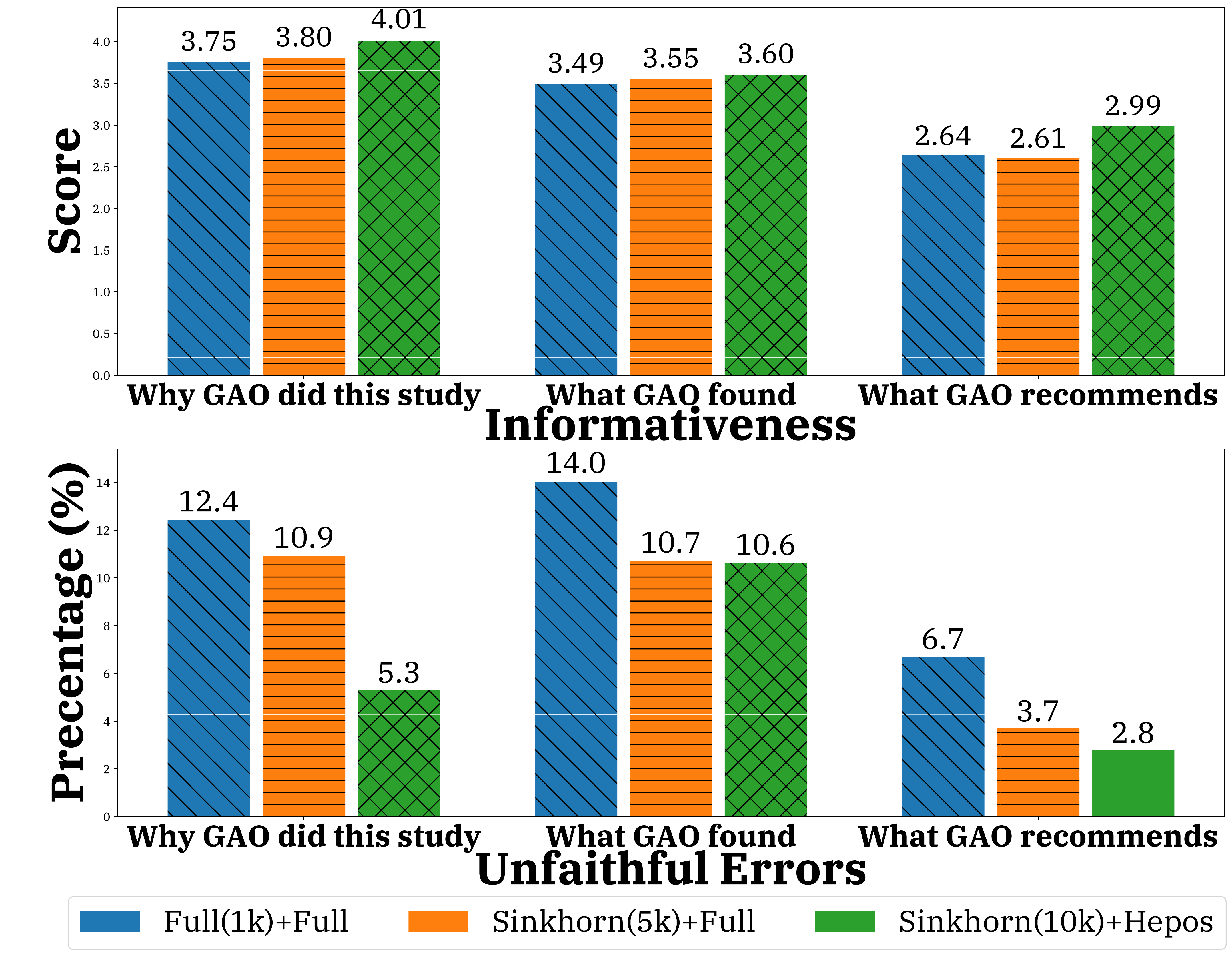}
    \vspace{-5mm}
    \captionof{figure}{Aspect-level informativeness and percentages of sentences with unfaithful errors on GovReport. }
    \label{fig:human_eval_aspect_level_gov}
\end{figure}

Next, we explore if reading more helps correctly reflect the content in documents' later sections. We plot aspect-level human ratings of informativeness and unfaithful errors on PubMed and GovReport in Fig.~\ref{fig:human_eval_aspect_level_pubmed} and Fig.~\ref{fig:human_eval_aspect_level_gov}. 
We report percentages of sentences with unfaithful errors by majority voting (i.e., at least one error is found by both annotators in the sentence). 
As can be seen, our models consistently improve informativeness and reduce errors across sections, especially for ``Results'' and ``Conclusions'' on PubMed and ``What GAO recommends'' on GovReport---these sections often appear in the later part of the source documents. 
Especially, we find that the full attention model tends to produce fabricated numbers in resultant summaries, whereas our models are able to correct them. 

\begin{table}[th]
    
    \fontsize{9}{11}\selectfont
    \setlength{\tabcolsep}{0.2mm}
    \hspace{-1mm}
    \begin{tabular}{lcccccc}
    \toprule
    {}  & \multicolumn{3}{c}{\bf GovReport} & \multicolumn{3}{c}{\bf PubMed}  \\
    System (MaxLen) & F. & APES & APES$_{src}$ & F. & APES & APES$_{src}$  \\
    \midrule
       \textsc{Full} (1024) & 58.9 & 42.7 & 42.7 & \textbf{74.6} & 43.2 & 31.5   \\
    \hline
    \multicolumn{7}{l}{\textbf{Encoder variants w/ full enc-dec attn.}} \\
    \textsc{Stride} (4096) & 55.3 & 43.1 & 42.5 & 72.7 & 43.8 & 31.9   \\
    \textsc{Lin.} (3072) & 48.4 & 35.7 & 36.3 & 67.7 & 39.3 & 29.5 \\
    \textsc{LSH} (4096) & 55.7 & \textbf{44.0} & 43.6 & 73.2 & 46.7 & 35.1   \\
    \textsc{Sinkhorn} (5120) & 57.0 & 43.6  & 42.1  & 72.9 & 46.8 & 35.4 \\
    \midrule
    \multicolumn{7}{l}{\textbf{Encoder variants w/ \textsc{Hepos} enc-dec attn.} (ours)} \\
    \textsc{LSH} (7168) & 59.6 & \textbf{44.0} & 44.2  & 73.3 & \textbf{47.5} & \textbf{35.6}  \\
    \textsc{Sinkhorn} (10240) & \textbf{60.1} & \textbf{44.0} & \textbf{44.3} & 71.9 & 46.2 & 34.8  \\

    \bottomrule
    \end{tabular}
    \caption{Evaluation with FactCC (F.), APES, and the new APES$_{src}$ metric, with higher numbers indicating more faithful summaries.}
    \label{tab:faithfulness}
\end{table}

\begin{table}[th]
    \centering
    \fontsize{9}{11}\selectfont
    \setlength{\tabcolsep}{2mm}
    \begin{tabular}{@{}lcccc@{}}
    \toprule
    {} & \multicolumn{2}{c}{\bf GovReport} & \multicolumn{2}{c}{\bf PubMed} \\
    Metric & Inf.$\uparrow$ & Err.$\downarrow$ & Inf.$\uparrow$ & Err.$\downarrow$ \\
    \midrule
    FactCC & 0.07 & -0.08 & 0.10 & -0.14  \\
    APES & 0.16 & -0.15 & 0.25 & -0.31  \\
    APES$_{src}$ & \textbf{0.21} & \textbf{-0.23}$\ast$ & \textbf{0.32}$\ast$ & \textbf{-0.32}  \\
    \bottomrule
    \end{tabular}
    \caption{
    Pearson correlation between human ratings and metrics. We use aggregated unfaithful errors (Err.). $\ast$: significantly better than other metrics based on William's test~\cite{williams1959regression} ($p<0.05$).
    }
    \label{tab:human_eval_correlation}
\end{table}

Lastly, we report the entailment-based FactCC and QA scores APES and APES$_{src}$ for top performing models in Table~\ref{tab:faithfulness}. 
The results again show that \textit{consuming longer input leads to more faithful summaries}, though the differences are less pronounced.

\subsection{Correlations between Human and Automatic Metrics} 
\label{subsec:autoeval}

Finally, we study whether the faithfulness evaluation metrics correlate with human judgment. 
As shown in Table~\ref{tab:human_eval_correlation}, on both government reports and scientific papers, \textit{QA metrics are better correlated with human ratings, with our newly proposed \textsc{APES$_{src}$} being the stronger of the two.}  After inspection, we find that human-written summaries contain paraphrases or acronyms that \textsc{APES} cannot capture via strict lexical matching. For instance, for the question ``Diabetes may worsen $\rule{0.6cm}{0.2mm}$ in patients'', the reference answer is ``death rate'', whereas answers from the source and the system summary are both ``mortality''. \textsc{APES$_{src}$} captures this, but not \textsc{APES}. 

\section{Additional Related Work}
\label{sec:add_related_work}

Summarizing long inputs has been investigated in many domains,  including books~\cite{mihalcea-ceylan-2007-explorations}, patents~\cite{trappey2009automatic}, movie scripts~\cite{gorinski-lapata-2015-movie}, and scientific publications~\cite{qazvinian2008scientific}. However, the datasets are often too small to train neural models.  \newcite{cohan-etal-2018-discourse} publish two large-scale datasets by collecting articles from \textsc{arXiv} and \textsc{PubMed}. Popular methods rely on extractive summarizers that identify salient sentences based on positional information~\cite{dong2020hiporank} or combined global and local contexts~\cite{xiao-carenini-2019-extractive}, where each sentence is represented as aggregated word embeddings. 
However, extractive summaries are often redundant and incoherent, highlighting the need for handling long documents via abstractive summarization.

To that end, extract-then-abstract methods are proposed. For example, \newcite{pilault-etal-2020-extractive} first extract relevant sentences and then rewrite them into paper abstracts. 
Our work is in line with building end-to-end abstractive summarization models for long input. \newcite{cohan-etal-2018-discourse} design a hierarchical encoder to read different sections separately, and then use combined attentions over words and sections to generate the summary. Multiple agents are created to read segments separately, and then collaboratively write an abstract~\cite{celikyilmaz2018deep}. However, both work truncates articles to $2$K words. 
Although efficient encoder attentions have been studied in \newcite{zaheer2020big} for abstractive summarization, at most $3$K tokens can be consumed by their models. Our \textsc{Hepos} encoder-decoder attention are able to process more than $10$K tokens, significantly improving summary informativeness and faithfulness. 
\section{Conclusion}
\label{sec:conclusion}

We investigate efficient attentions for long document summarization. We propose a novel encoder-decoder attention, \textsc{Hepos}, based on head-wise positional strides that can effectively identify salient content. Models based on \textsc{Hepos} attention can process at least twice as many words and produce more informative summaries with less unfaithful errors, according to both automatic evaluation and human evaluation. We further show that our new cloze QA metric better correlates with human judgment than prior faithfulness evaluation metrics.

\section*{Acknowledgements}
This research is supported in part by Oracle for Research Cloud Credits, National Science Foundation through Grant IIS-1813341, and by the Office of the Director of National Intelligence (ODNI), Intelligence Advanced Research Projects Activity (IARPA), via contract \# FA8650-17-C-9116. The views and conclusions contained herein are those of the authors and should not be interpreted as necessarily representing the official policies, either expressed or implied, of ODNI, IARPA, or the U.S. Government. The U.S. Government is authorized to reproduce and distribute reprints for governmental purposes notwithstanding any copyright annotation therein. 
We thank three anonymous reviewers for their valuable suggestions and comments.

\newpage
\bibliography{anthology,custom}
\bibliographystyle{acl_natbib}

\appendix

\section{GovReport Dataset Collection and Processing}\label{sec:append_gov_dataset}

For GAO reports, their summaries are organized as highlights. We collect GAO reports that include corresponding highlights and were published before Jul 7, 2020 . The reports and highlights are published in PDF files. Most of the highlights are also reorganized and shown on the web page as HTML. Since PDF parsing is more prone to errors than web parsing, we only keep the reports whose highlights can be obtained on the corresponding web page to ensure the quality of extracted gold-standard summaries. For reports, we first convert the PDF files to HTML using PDFMiner\footnote{\url{https://github.com/euske/pdfminer}}. We then parse the HTML into text into sections and paragraphs with handcrafted parsing rules. 
We remove the reports that do not have cover pages, as our rules are constructed for documents with then. 
We further remove parsed documents with empty sections, non-capitalized section titles, or a single section, since these are common patterns of incorrectly parsed documents. Failed parsing would also result in short documents. Therefore, we examine the reports with shorter length and then filter out $10\%$ of the shortest reports. 

We collect CRS reports that were published before May 20, 2020 from EveryCRSReport\footnote{\url{https://www.everycrsreport.com}} where the original PDF files are already parsed into HTML. We only keep documents with expert-written summaries. We then gather texts from the html files.

\section{Experiment Details}\label{sec:append_eval_detail}

\noindent \textbf{FactCC Training Data Construction}. \newcite{kryscinski-etal-2020-evaluating} generate training data by applying rule-based transformations to sentences from source documents. 
We leverage reference summaries, where we train a FactCC model by reading a summary sentence (i.e., the claim) and a context to predict the corresponding label. 
A context is constructed by greedily selecting sentences that maximize the improvement of its ROUGE-2 when compared against the reference summary sentence. Following FactCC, we apply \textit{sentence negation}, \textit{entity swap}, and \textit{number swap} to summary sentences to construct negative claims and use the original sentences as positive claims. During testing, we first find the context for each system summary sentence. The model then predicts a sentence-level faithfulness score by reading the system summary sentence and the context. 

\smallskip 
\noindent \textbf{Evaluation Model Training}. We fine-tune BERT~\cite{devlin-etal-2019-bert} for both FactCC and QA models. We include an additional classification head to predict entailment label or answer spans based on the [CLS] token. For GovReport dataset, we consider a base version of BERT with uncased tokens. For PubMed, we use a BERT model which is fine-tuned on PubMed abstracts to obtain better performance\footnote{\url{https://huggingface.co/monologg/biobert\_v1.0\_pubmed\_pmc}}. 

\smallskip 
\noindent \textbf{Entity Extraction Model}. We use OneIE to extract entities from the reference summary~\cite{lin2020oneie}. OneIE is a unified framework that combines entities, relations, and events extraction in one model. The model leverages the BERT pre-trained weights as the sentence embedding to produce entities, relations, and events from a sentence. Two OneIE models are built.

The first model for government reports is trained on the Automatic Content Extraction (ACE) 2005 dataset~\cite{ace2005}. This model can extract entities from general conversation contexts such as \textit{People}, \textit{Location}, or \textit{Organization}, and events such as \textit{Movement}, \textit{Conflict}, or \textit{Justice}, etc.

The second model for scientific domain information extraction is trained on the Genia 2011~\cite{genia2011}, Genia 2013 \cite{genia2013}, and PubMed~\cite{wei2019pubtator} datasets. It extracts entity such as \textit{Gene}, \textit{Variant}, \textit{Disease}, \textit{Chemical}, or \textit{Species}, and events such as \textit{Gene Expression}, \textit{Binding}, \textit{Protein Modification}, or \textit{Positive Regulation}, etc. The full list of entity and event types can be found in Table~\ref{tab:biooneie-dataset}. To train this model, we fine-tune the BioBERT pre-trained model~\cite{lee2020biobert} on the COVID-19 Open Research (CORD-19) dataset~\cite{wang2020cord}. As we proposed, this model is applied to the PubMed data.

\begin{table}[t]
    \centering
    \fontsize{9}{11}\selectfont
    \setlength{\tabcolsep}{0.3mm}
    \begin{tabular}{@{}lrrr@{}}
    \toprule
& \textbf{Genia 2011}  & \textbf{Genia 2013}  & \textbf{PubMed}  \\ 
    \hline
\multicolumn{4}{l}{\textbf{Entity Type}} \\
Anaphora & -  & 105  & -  \\ 
Entity & 480  & 121  & -  \\ 
CellLine & -  & -  & 614  \\ 
Chemical & -  & -  & 14,051  \\ 
Disease & -  & -  & 62,228  \\ 
Mutation & -  & -  & 164  \\ 
Protein & 11,539  & 3,562  & 15,577  \\ 
Species & -  & -  & 52,954  \\ 
\multicolumn{4}{l}{\textbf{Event Type}} \\
Binding & 880  & 167  & -  \\ 
Gene Expression & 2,076  & 666  & -  \\ 
Localization & 264  & 44  & -  \\ 
Negative Regulation & 338  & 273  & -  \\ 
Phosphorylation & 175  & 105  & -  \\ 
Positive Regulation & 1,123  & 311  & -  \\ 
Protein Catabolism & 100  & 23  & -  \\ 
Protein Modification & -  & 8  & -  \\ 
Regulation & 292  & 72  & -  \\ 
Transcription & 580  & 97  & -  \\ 
Ubiquitination & -  & 4  & -  \\ 
    \bottomrule
    \end{tabular}
    \caption{Dataset description for training OneIE Biomedical extraction. While Genia 2011 and 2013 datasets focus more on event extraction, PubMed covers more entities.}
    \vspace{-3mm}
    \label{tab:biooneie-dataset}
\end{table}

\section{Additional Sample Outputs}\label{sec:append_sample_output}

We include two samples from GovReport and PubMed to further illustrate that our model with \textsc{Hepos} attention generates more faithful and informative summaries in Fig.~\ref{fig:sample_outputs_gov} and Fig.~\ref{fig:sample_outputs_pubmed}.

\section{Human Evaluation Guideline}\label{sec:append_human_eval}

In human evaluation, annotators are asked to evaluate the system summaries generated for a report or a paper. In addition to the summaries, annotators are provided with \textit{the report or the paper to be summarized} and a corresponding \textit{human-written reference}. 
Human judges evaluate each system summary \textbf{sentence by sentence}. The annotation consists of three tasks, which are described below. 

\textbf{Task 1: Aspect Labeling}. First, annotators are asked to decide which \textbf{aspect} each sentence belongs to. For government reports, each sentence should be categorized into three aspects: (1) Why GAO did this study, (2) What GAO found, and (3) What GAO recommends. 
For scientific papers, summaries have four aspects: (1) Introduction and Literature, (2) Methods, (3) Results, and (4) Discussion and Conclusion. Table~\ref{table:aspect} and Table~\ref{table:aspect_pubmed} contain example reference summaries with labeled aspects.

\textbf{Task 2: Sentence-level Faithfulness Error Labeling}. 
Next, annotators will judge whether each sentence contains any \textbf{unfaithful content}. {Unfaithful} content is categorized into three types. A ``0'' or ``1'' label will be given to each type, where ``0'' indicates the sentence is free of such type of error, and ``1'' otherwise.  

Concretely, {unfaithful content} is the fabricated or contradictory content which \textit{is not present or contradicts the facts} in the source article. It can also be \textit{ambiguous expression} which distorts the meaning. Here are detailed descriptions for the three types of errors: 

\begin{itemize}
    \item \textbf{Hallucination} error refers to fabricated content that cannot be found or inferred from the source.
    \item Misconstruction error that is due to \textbf{deletion} of entities, events, or clauses, resulting in sentences that are incomplete, missing context, or ungrammatical.
    \item
    Misconstruction error that is caused by \textbf{false concatenation} of content from different places in the source. 
\end{itemize}

\textbf{Task 3: Aspect-level Summary Quality Rating}. 
After reading the full summary, annotators will evaluate the 
\textbf{informativeness} of the summary for each aspect--- whether the summary provides \textit{a necessary and enough coverage of information} in the \textit{reference}. For instance, whether the summary covers all the salient points in ``Why GAO did this study".

Here are detailed descriptions of {informativeness}: 
\begin{itemize}
    \item \textbf{5}: Summary covers enough key points in the reference (only misses minor topics), and is free of unfaithful errors.
    
    \item \textbf{4}:
    Summary covers major key points (e.g., 80 percent) and may miss one or two key points in the reference. Summary can contain one unfaithful error. 
    \item \textbf{3}:
    Summary covers roughly half of the key points in the reference or contains 2 or 3 unfaithful errors. 

    \item \textbf{2}:
    Summary only covers 1 or 2 key points and misses many important topics (e.g. > 80 percent) in the reference, or contains more than 3 major unfaithful errors, e.g. summary fabricates or distorts some facts. 

    \item \textbf{1}:
    Summary is irrelevant and does not cover any content in the reference.
\end{itemize}

\begin{figure*}
    \centering
    \fontsize{10}{12}\selectfont
    \setlength{\tabcolsep}{1.5mm}{
    \begin{tabular}{p{150mm}}
    \toprule
    \textbf{Summary by Human:} \\
    \textit{What GAO found} \\
    The Food and Drug Administration (FDA) and U.S. Department of Agriculture (USDA) have begun \hlc[blue!25]{collaborating on regulatory oversight} of cell-cultured meat. For example, in 2019, the agencies \hlc[blue!25]{signed an interagency agreement} and \hlc[blue!25]{created three working groups} \hlc[blue!25]{to carry out the terms of the agreement.} However, the agreement and working groups could more fully incorporate practices to enhance and sustain collaboration, such as defining outcomes. For example, the agreement identifies the development of labeling principles as an outcome, but does not describe how the agencies will track and monitor progress toward this outcome, and the working groups identify a lead agency but not members' roles. (...) \\
    \hlc[blue!25]{Developing and updating written guidance and agreements} is also a leading practice for interagency collaboration. By fully incorporating leading practices into their efforts to collaborate, the agencies could minimize potential overlap and fragmentation, use resources in a more efficient manner, and better ensure the public and other key stakeholders have clarity about the agencies' oversight responsibilities. (...) \\
    \hline
    \textbf{Model w/ full attn.:} \\
    \textit{What GAO found} \\
    The frequency of inspections the agencies conduct varies, \hlc[red!25]{as follows: FDA}. FDA conducts risk-based inspections of high- risk and non-high-risk food facilities. FDA's authority requires a risk- based approach, in which inspection rates vary depending on the level of risk associated with a food product. (...) \\
    In 2010, \hlc[red!25]{FDA began conducting inspections of meat slaughter facilities} to determine whether they were complying with the agency's longstanding interpretation of its statutes requiring it to do so. In addition,\hlc[red!25]{USDA conducted inspections of slaughter facilities} that process catfish to determine if they are complying with USDA's requirement to inspect catfish slaughter facilities. USDA also conducts inspections of processing facilities to verify that they are following USDA's requirements to inspect these facilities. In 2011, USDA began inspecting processing facilities for catfish for the first time. (...) \\
    \hline
    \textbf{Model w/ HEPOS enc-dec attn.:} \\
    \textit{What GAO found} \\
    The Departments of Health and Human Services (HHS) and Agriculture have taken several steps to \hlc[blue!25]{coordinate their oversight of the development}, commercialization, and regulation of cell cultured meat and seafood, but they have not fully incorporated leading practices for interagency collaboration. For example, HHS and USDA have \hlc[blue!25]{established interagency working groups} that have met to discuss issues such as coordination and coordination. However, the agencies have not yet fully incorporated some leading practices, such as \hlc[blue!25]{establishing clear roles and responsibilities, into their interagency agreements.} (...) \\
    \bottomrule
    \end{tabular}
    }
    \caption{Sample summaries for a government report. Model with truncated input generates \hlc[red!25]{unfaithful content}. Our HEPOS encoder-decoder attention with Sinkhorn encoder attention covers more \hlc[blue!25]{salient information} in ``What GAO found'' aspect.}
    \label{fig:sample_outputs_gov}
\end{figure*}

\begin{figure*}
    \centering
    \fontsize{10}{11}\selectfont
    \setlength{\tabcolsep}{1.5mm}{
    \begin{tabular}{p{150mm}}
    \toprule
     \textbf{Original Paper:} \\
     (...) anatomical abnormalities included complex pelvicalyceal anatomy in 98, pelviureteric junction obstruction (pujo) in 35, ectopic, horseshoe, cross-fused ectopic kidneys in 40, and duplex system with nonfunctioning upper or lower moiety in 14 patients. (...) \\
     of the 1404 patients from rural area , 547 (39\%) were managed by open surgery as compared to 362 (22\%) of the patients from urban areas. (...) \\
     the mean size of the kidney stone was 5.05 5.88 cm, ureter 1.95 1.33 cm, and bladder 7.6 3.2 cm. (...)\\
     overall success rate was 90\% in pcnl, 96\% in urs, 100\% in cystolithotripsy, and 81\% in swl. \\
    \midrule
    \textbf{Summary by Human:} \\
    \textit{Results} \\
    Results of 3969 surgeries, 2794 (70\%) were minimally invasive surgery (mis) techniques to include swl 19\%, pcnl 16 \% , urs 18.9\% , and pucl + pccl 16\% and 1175 (30\%) were open surgeries . \\
    the main factors necessitating open surgery were large stone burden 37\%, anatomical abnormalities 16\%, stones with renal failure 34\%, gross hydronephrosis with thin cortex 58\% , urinary tract infection (uti) 25\%, and failed mis 18\%. nearly 50\% of the surgeries were necessitated by economic constraints and long distance from center where one-time treatment was preferred by the patient.
    \hlc[blue!25]{tone-free rates by open surgeries were pyelolithotomy 91\%}, ureterolithotomy 100\%, and cystolithotomy 100\% with complication rate of up to 3\%. \\
    \hline
    \textbf{Model w/ full attn.:} \\
    \textit{Results} \\
    results: the mean age of the patients was 6.5 years. the mean follow-up period was 12.5 months. mean stone size was 1.5cm. the majority of the stones were lager stones (n=454). \hlc[red!25]{the stone free rate was 100\%.} open surgery was performed in children with large bulk of stones, anatomical abnormalities, marked obstructive cortical atrophy and scarring, gross hydronephrosis, or urinary tract infection (uti).
     \\
    \hline
    \textbf{Model w/ HEPOS enc-dec attn.:} \\
    \textit{Results} \\
    the mean age of the patients was 6.35, 3.7 years with a male-to-female ratio of 2.8:1, and a range of 25 days (<1 month) to 15 years. anatomical abnormalities included complex pelvicalyceal anatomy in 98, pelviureteric junction obstruction (pujo) in 35, ectopic, horseshoe, cross-fused ectopic kidneys in 40, and duplex system with nonfunctioning upper or lower moiety in 14 patients. the mean size of the kidney stone was 5.05 5.88 cm3. of the 1404 patients from rural areas, 547 (39\%) were managed by surgery as compared to 362 (22\%) patients from urban areas. \hlc[blue!25]{overall success rate was 90\% in pcnl} , 96\% in urs , 100\% in cystolithotripsy , and 81\% in swl. \\
    \bottomrule
    \end{tabular}
    }
    \caption{Sample summaries for a scientific paper. Model with truncated input generates \hlc[red!25]{fabricated facts}. Our HEPOS encoder-decoder attention with LSH encoder attention are more \hlc[blue!25]{faithful} for the aspect of ``results''.}
    \label{fig:sample_outputs_pubmed}
\end{figure*}
\begin{table*}
    \centering
    \fontsize{10}{11}\selectfont
    \begin{tabularx}{\textwidth}{|l|X|}
    \toprule
        \textbf{Aspect} & \textbf{Example}  \\
\midrule
      Why GAO Did This Study  &  To protect data that are shared with state government agencies, federal agencies have established cybersecurity requirements and related compliance assessment programs.  Specifically, they have numerous cybersecurity requirements for states to follow when accessing, storing, and transmitting federal data. GAO was asked to evaluate federal agencies' cybersecurity requirements and related assessment programs for state agencies. The objectives were to determine the extent to which (...) \\
      \midrule
      What GAO Found   & Although the Centers for Medicare and Medicaid Services (CMS), Federal Bureau of Investigation (FBI), Internal Revenue Service (IRS), and Social Security Administration (SSA) each established requirements to secure data that states receive, these requirements often had conflicting parameters. Such parameters involve agencies defining specific values like the number of consecutive unsuccessful logon attempts prior to locking out the user. Among the four federal agencies, the percentage of total requirements with conflicting parameters ranged from 49 percent to 79 percent. Regarding variance with National Institute of Standards and Technology guidance, GAO found that the extent to which the four agencies did not fully address guidance varied from 9 percent to 53 percent of total requirements. The variances were due in part to the federal agencies' insufficient coordination in establishing requirements. (...) \\
      
      \midrule
      What GAO Recommends  & GAO is making 12 recommendations to the four selected agencies and to OMB. Three agencies agreed with the recommendations and one agency (IRS) partially agreed or disagreed with them. OMB did not provide comments. GAO continues to believe all recommendations are warranted. \\
\bottomrule
    \end{tabularx}
    \caption{Sample reference summary with aspects in a GAO report.}
    \label{table:aspect}
\end{table*}

\begin{table*}
    \centering
    \fontsize{10}{11}\selectfont
    \begin{tabularx}{\textwidth}{|l|X|X|}
    \toprule
        \textbf{Aspect} & \textbf{Keywords} & \textbf{Example}  \\
\midrule
      Introduction and Literature  &  introduction, case, objectives, purposes, objective, purpose, background, literature, related work & background : the present study was carried out to assess the effects of community nutrition intervention based on advocacy approach on malnutrition status among school - aged children in shiraz , iran . \\
      && \\
      && introduction . low serum vitamin d levels are associated with increased postural sway . vitamin d varies seasonally . this study investigates whether postural sway varies seasonally and is associated with serum vitamin d and falls . \\
      \midrule
      Methods  & materials and methods, techniques, methodology, materials, research design, study design & materials and methods : this case - control nutritional intervention has been done between 2008 and 2009 on 2897 primary and secondary school boys and girls ( 7 - 13 years old ) based on advocacy approach in shiraz , iran . the project provided nutritious snacks in public schools over a 2 - year period along with advocacy oriented actions in order to implement and promote nutritional intervention . for evaluation of effectiveness of the intervention growth monitoring indices of pre- and post - intervention were statistically compared . \\
      \midrule
      Results  & results, experiments, observations & results : the frequency of subjects with body mass index lower than 5\% decreased significantly after intervention among girls ( p = 0. 02 ) . however , there were no significant changes among boys or total population . (...) \\
      \midrule
      Discussion and Conlusion  & discussion, limitation, conclusions, concluding &  conclusion : this study demonstrates the potential success and scalability of school feeding programs in iran . community nutrition intervention based on the advocacy process model is effective on reducing the prevalence of underweight specifically among female school aged children . \\

\bottomrule
    \end{tabularx}
    \caption{Sample reference summary with aspects labeled in a PubMed article. Keywords are used to match different parts of the summaries to the four aspects. 
    }
    \label{table:aspect_pubmed}
\end{table*}


\end{document}